\documentclass{article}

    \PassOptionsToPackage{numbers, compress}{natbib}
    
\usepackage[preprint]{neurips_2024}

\usepackage[utf8]{inputenc} 
\usepackage[T1]{fontenc}    
\usepackage{url}            
\usepackage{booktabs}       
\usepackage{amsfonts}       
\usepackage{nicefrac}       
\usepackage{microtype}      
\usepackage{xcolor}         

\usepackage{microtype}
\usepackage{graphicx}
\usepackage{floatrow}
\usepackage{subcaption}
\usepackage{listings}
\usepackage{longtable}

\usepackage[bookmarks=true,         colorlinks=true,linkcolor=red,urlcolor=magenta,citecolor=green!80!black]{hyperref}

\usepackage{multirow}
\usepackage{color}
\definecolor{deepblue}{rgb}{0,0,0.5}
\definecolor{deepred}{rgb}{0.6,0,0}
\definecolor{deepgreen}{rgb}{0,0.5,0}

\usepackage{amsmath}
\usepackage{amssymb}
\usepackage{mathtools}
\usepackage{amsthm}
\theoremstyle{plain}

\theoremstyle{definition}

\theoremstyle{remark}

\usepackage{bm} 
\usepackage{multirow}
\usepackage{algorithm, algorithmic}
\usepackage[nameinlink,capitalise]{cleveref}
\crefname{section}{§}{§§}
\Crefname{section}{§}{§§}
\crefname{lemma}{lemma}{lemmas}
\Crefname{lemma}{Lemma}{Lemmas}
\crefname{thm}{theorem}{theorems}
\Crefname{thm}{Theorem}{Theorems}

\usepackage[numbers]{natbib}
\title{Learnable Evolutionary Multi-Objective Combinatorial Optimization via Sequence-to-Sequence Model}

\author{
Jiaxiang Huang\textsuperscript{1}, Licheng Jiao\textsuperscript{1}\thanks{Correspondence to: Licheng Jiao <lchjiao@mail.xidian.edu.cn>.}\\[0.1cm]
\textsuperscript{1}Key Laboratory of Intelligent Perception and Image Understanding of the Ministry of\\ Education of China, School of Artificial Intelligence, Xidian University\\[0.1cm]
\texttt{jxhuang@stu.xidian.edu.cn, lchjiao@mail.xidian.edu.cn} \\[0.3cm]
}

\begin{document}

\maketitle

\begin{abstract}
  Recent advances in learnable evolutionary algorithms have demonstrated the importance of leveraging population distribution information and historical evolutionary trajectories. While significant progress has been made in continuous optimization domains, combinatorial optimization problems remain challenging due to their discrete nature and complex solution spaces. To address this gap, we propose SeqMO, a novel learnable multi-objective combinatorial optimization method that integrates sequence-to-sequence models with evolutionary algorithms. Our approach divides approximate Pareto solution sets based on their objective values' distance to the Pareto front, and establishes mapping relationships between solutions by minimizing objective vector angles in the target space. This mapping creates structured training data for pointer networks, which learns to predict promising solution trajectories in the discrete search space. The trained model then guides the evolutionary process by generating new candidate solutions while maintaining population diversity. Experiments on the multi-objective travel salesman problem and the multi-objective quadratic assignment problem verify the effectiveness of the algorithm. Our code is available at: \href{https://github.com/jiaxianghuang/SeqMO}{https://github.com/jiaxianghuang/SeqMO}.
\end{abstract}
\section{Introduction}\label{sec: introduction}

Combinatorial optimization problems often face combinatorial explosion due to their high-dimensional solution space. Metaheuristic methods, which do not require analytical objective functions or gradient information and are widely applicable to various problems, are commonly used to solve combinatorial optimization problems. For more challenging Multi-Objective Combinatorial Optimization (MOCO) problems, evolutionary multi-objective optimization methods are frequently employed. These methods perform genetic operations, such as crossover and mutation operators, on populations composed of multiple solutions, to generate new offspring and iteratively update the previous population. Current evolutionary multi-objective combinatorial optimization methods rarely utilize problem information or historical solution information from the evolutionary process to fully explore and exploit better solutions. Particularly for permutation-based combinatorial optimization problems, the challenge is heightened because solutions must satisfy permutation constraints, and changes in several dimensions of a permutation may lead to drastic changes in objective function values.

With the rapid development of deep learning technology across various fields, numerous achievements have validated the effectiveness of data-driven learning methods \cite{he2020evolutionary,liu2022learning,jin2018data,liu2023survey,zhang2011evolutionary}. Learnable evolutionary optimization, which combining the traditional evolutionary optimization with deep learning techniques, is becoming a hot research topic in the optimization area. However, the learnable evolutionary optimization methods are mainly forcus on the continuous optimization problems. The combinatorial optimization  present more intrinsic challenges due to their discrete nature. 

In this paper, focusing on permutation-based combinatorial optimization problems, we propose a learnable evolutionary multi-objective combinatorial optimization method based on a sequence-to-sequence model. This method fully utilizes the historical information of solution updates from the multi-objective evolutionary optimization process. Through training and learning the historical iteration process of solutions using a sequence-to-sequence model, it predicts new Pareto approximate solutions. Specifically, the method classifies Pareto approximate solutions generated in each generation, using high-quality solutions near the Pareto front as training labels and poor-quality solutions far from the Pareto front as training data. Through the relative superiority in objective space, it predicts possible movement trajectories of solutions in solution space. By iteratively training old trajectories and predicting new ones, the method achieves approximation to the Pareto front. The algorithm's effectiveness is verified on multi-objective traveling salesman problems and quadratic assignment problems.

\section{SeqMO}\label{sec:seqmo}

\begin{figure}[htbp]
	\centering
	\includegraphics[width=3.0in]{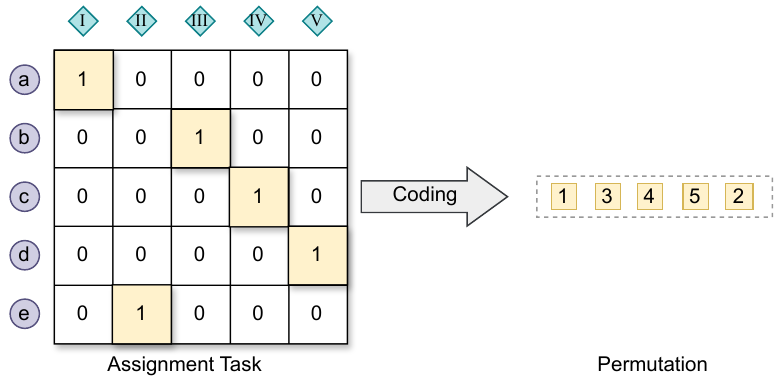}
	\centering
	\caption{Permutation-Based Combinatorial Optimization}
	\label{fig:permutation}
\end{figure}

Permutation-based combinatorial optimization aims to find an optimal permutation of the sequence $1,2,\dots,N$ to maximize objective functions, such as in task assignment problems and matching problems. As shown in Figure~\ref{fig:permutation}, an Assignment Task $T$ can be abstractly encoded as a permutation $[1,3,4,5,2]$, where $1$ in the permutation indicates that person a is assigned task I, $3$ indicates that person $b$ is assigned task III, and so on. Generally, permutation combinatorial optimization problems must satisfy the following constraints:

\begin{equation}
	\left\{\begin{matrix}
		T_{ij}\in \{0,1\}\\
		i\in \{1,2,\cdots ,N\}\\
		j\in \{1,2,\cdots ,N\}\\	
		\sum_{i=1}^{N} T_{ij}=1, \forall j\\
		\sum_{j=1}^{N} T_{ij}=1, \forall i\\
	\end{matrix}\right.
	\label{eq:permutation}
\end{equation}

Permutation-based combinatorial optimization problems can typically be solved using integer programming \cite{hahn1998branch}, spectral relaxation \cite{leordeanu2005spectral}, heuristic methods \cite{rosenkrantz1977analysis}, and metaheuristic methods \cite{grefenstette2014genetic,saraph2014magna}. In recent years, with the development of deep learning technology, methods combining machine learning with traditional approaches have rapidly evolved. While metaheuristic methods provide a general framework with broad applicability across various problems, they often fail to fully utilize historical and currently generated data. Combining them with machine learning-based methods has proven more effective for solving permutation-based combinatorial optimization problems.

As shown in equation (\ref{eq:permutation}), when solving permutation-based combinatorial optimization problems by metaheuristic methods, encoding the solutions of the problems to satisfy constraints is crucial. Real-number encoding facilitates both population generation and updating in metaheuristic methods, as well as gradient backpropagation in machine learning. However, it often struggles to satisfy permutation constraints, resulting in numerous invalid solutions. Here, population generated by metaheuristic methods is used to train an external Pointer Network \cite{vinyals2015pointer}, which in turn can update solutions in the metaheuristic method itself, creating an iterative evolutionary loop. This approach effectively combines the advantages of machine learning and metaheuristic methods to solve permutation-based combinatorial optimization problems.

\begin{figure}[htbp]
	\centering
	\includegraphics[width=4.5in]{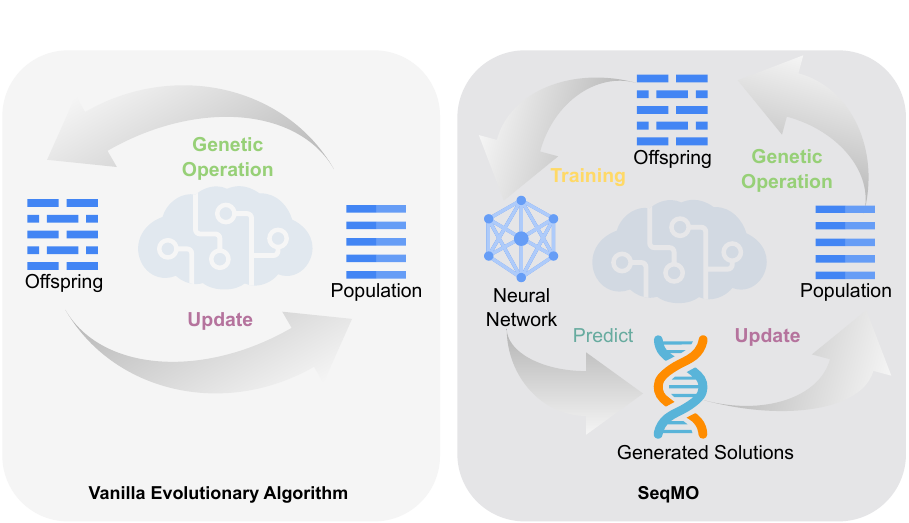}
	\centering
	\caption{Vanilla Evolutionary Algorithm VS. SeqMO}
	\label{fig:trainevo}
	\end{figure}
	
The algorithm combines the advantages of neural network learning and evolutionary multi-objective optimization to enhance the optimization process. As shown in Figure~\ref{fig:trainevo}, evolutionary optimization continuously generates offsprings, while the neural network utilizes non-dominance relationships or decomposed multi-objective optimization sub-problem objective values to construct training data and labels. These data and labels are used to train a neural network, such as a Pointer Network. During the prediction phase, the trained network generates new offsprings, which enhance the exploration and exploitation capabilities of the original offsprings. In the update phase, the original offsprings and new offsprings are used to update the population. Through this iterative cycle, the next generation's population generates new offsprings that can serve as training data, continuously updating the training data and thereby updating the neural network parameters.

\begin{algorithm}[htbp]
	\caption{SeqMO}
	\label{alg:trainevo}
	\begin{algorithmic}[1]
	\STATE \textbf{Input:}
	\STATE Population size: \emph{N}
	\STATE Maximal number of objective function evaluations: \emph{MaxFE}
	\STATE
	\STATE \textbf{1) Initialization:}
	\STATE $P \gets $ Initialize()
	\STATE
	\STATE \textbf{2) Optimization:}
	\WHILE{$FE \leq MaxFE$}
		\STATE \textbf{a) Genetic Operation:}
		\STATE $C \gets \varnothing$
		\FOR{$i \gets 1$ \TO \emph{N}}
			\STATE Select parents $u \in P$ and $v \in P$
			\STATE $c \gets $ GeneticOperator$(u, v)$
			\STATE $C \gets C \cup \{c\}$
		\ENDFOR
		
		\STATE \textbf{b) Dividing:}
		\STATE $(C_{\text{poor}}, C_{\text{elite}}) \gets$ NondominatedSorting($C$)
		
		\STATE \textbf{c) Constructing Data and Label:}
		\STATE (Option 1: Greedy)
		\STATE $\left\langle Data, Label \right\rangle \gets \varnothing$
		\FOR{each $p \in C_{\text{poor}}$}
			\STATE $e_{\text{min}} \gets \underset{e}{\operatorname{arg\,min}}\ \arccos(f(p), f(e)),\ \forall e \in C_{\text{elite}}$
			\STATE $\left\langle Data, Label \right\rangle \gets \left\langle Data, Label \right\rangle \cup \{(p, e_{\text{min}})\}$
		\ENDFOR
		
		\STATE (Option 2: Hungarian)
		\STATE $\measuredangle \gets \arccos(f(p), f(e)),\ \forall p \in C_{\text{poor}},\ \forall e \in C_{\text{elite}}$
		\STATE $\left\langle Data, Label \right\rangle \gets$ Hungarian($\measuredangle$)
		
		\STATE \textbf{d) Training:}
		\STATE $\theta \gets \varnothing$
		\STATE $\theta \gets$ PointerNetwork($\left\langle Data, Label \right\rangle$)
		
		\STATE \textbf{e) Predict:}
		\STATE $C_{\text{generated}} \gets$ PointerNetwork($\left\langle Data \right\rangle, \theta$)
		
		\STATE \textbf{f) Update:}
		\STATE $P \gets$ EnvironmentSelection($P, C, C_{\text{generated}}$)
	\ENDWHILE
	\STATE
	\STATE \textbf{Output:}
	\STATE Solution set to MOCO problem: population $P$
	\end{algorithmic}
\end{algorithm}

Algorithm \ref{alg:trainevo} details the population training-based multi-objective optimization algorithm. The iteration process includes genetic operators, offsprings division, dataset and label construction, Pointer Network training and prediction, and population update steps. The following sections primarily introduce the dataset and label construction, and the Pointer Network training and prediction components.

\begin{figure}[htbp]
\centering
\includegraphics[width=5.5in]{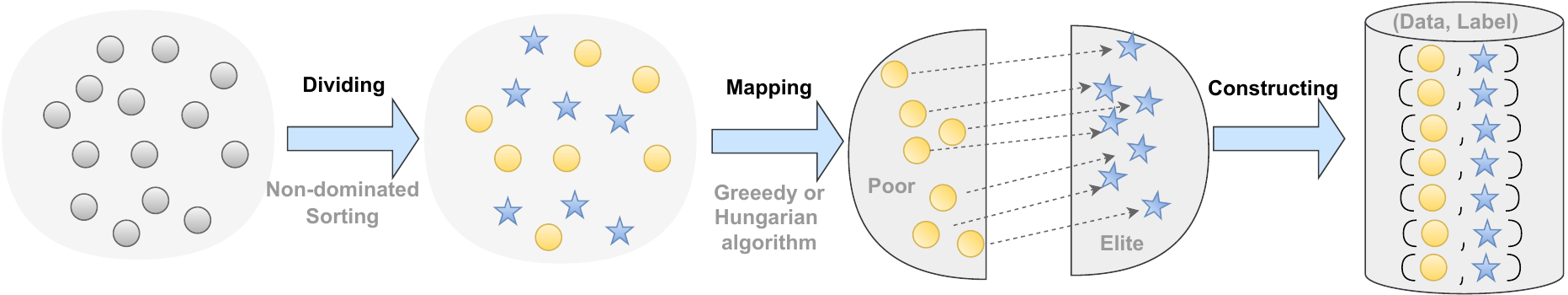}
\centering
\caption{Construction of Training Data and Labels}
\label{fig:construct}
\end{figure}

Training using historical population data from the evolutionary process, where the construction of training data and labels significantly affects the performance of the trained neural network. Since offsprings are newly generated each time, they offer better diversity and can provide more balanced training data. The construction of training data and labels involves dividing the data into training data and training labels, and establishing one-to-one matching between each data point and label. Figure \ref{fig:construct} illustrates the process of constructing training sets and labels.

During the division process, non-dominated sorting methods can be used to rank solutions, or comparisons can be made using decomposed and weighted sub-problem objective values. The divided offsprings is separated into two parts based on non-dominated sorting: elite solutions $C_{elite}$ closer to the Pareto front, and poor solutions $C_{poor}$ farther from the Pareto front. When establishing one-to-one matching, if the angle between a poor solution's objective function vector and an elite solution's objective function vector in objective space is smaller than that of other pairs, they can form a matching pair as training data and label.

 A greedy method can be used to find these data-label pairs\cite{liu2022learning}. For each poor solution $p \in C_{poor}$, find the elite solution $e_{min}$, whose objective vector has the minimum angle with $p$'s objective vector $\mathbf{f}(p)$, forming a training data-label pair$\left \langle p, e_{min} \right \rangle$, The calculation is as follows:

\begin{equation}
	\left\{\begin{matrix}
		e_{min} = \underset{e \in C_{elite}}{\operatorname{arg\,min}} \arccos(\mathbf{f}(p),\mathbf{f}(e))\\
		\arccos(\mathbf{u},\mathbf{v}) = \arccos \left( \frac{\mathbf{u} \cdot \mathbf{v}}{\|\mathbf{u}\| \|\mathbf{v}\|} \right) 
		\\
	\end{matrix}\right.
	\label{eq:arccos}
\end{equation}

The greedy algorithm may result in uneven angles between solution pairs. The Hungarian algorithm achieves one-to-one matching by minimizing overall loss, calculated as:
\begin{equation}
	\left \langle Data, Label \right \rangle =  Hungarian(\measuredangle(\mathbf{f}(C_{poor}),\mathbf{f}(C_{elite})))
\end{equation}
where Data represents all training data, $Label$ represents all labels; $\measuredangle$ represents the angle matrix or negative cosine value matrix between $C_{poor}$ and $C_{elite}$.

\begin{figure}[htbp]
\centering
\includegraphics[width=5.5in]{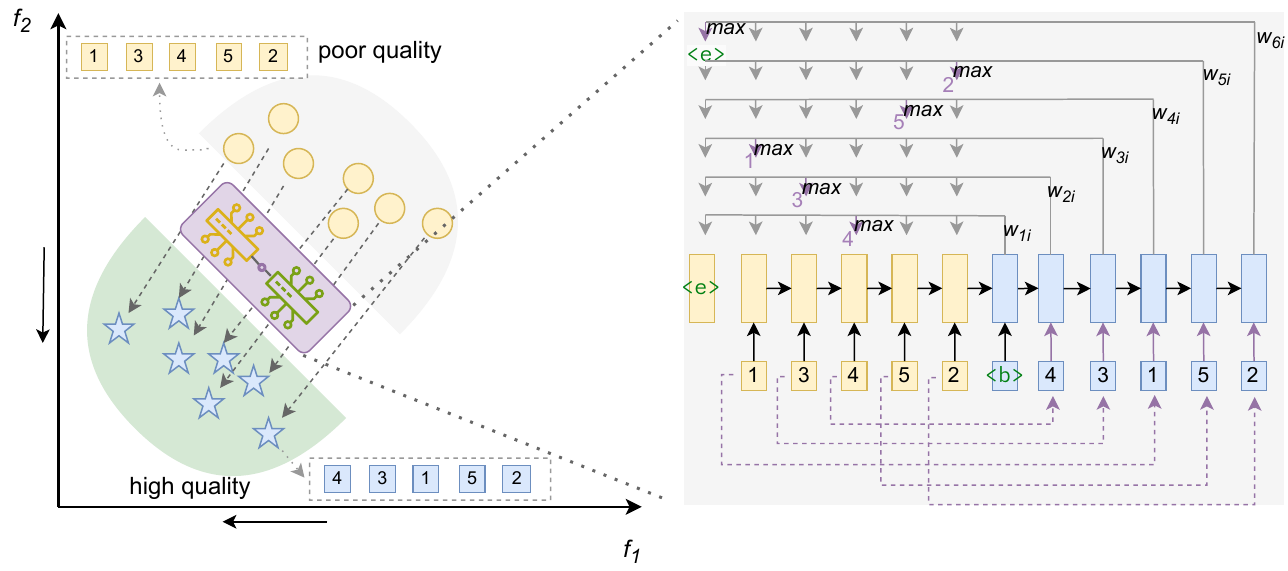}
\centering
\caption{Solution Generation Based on Pointer Networks}
\label{fig:mapping}
\end{figure}

As shown in Figure~\ref{fig:mapping}, the neural network aims to establish mapping relationships from poor solutions to elite solutions through training data constructed from elite and poor solutions. During prediction, poor solutions are migrated toward elite solution objective regions to discover higher-quality or more diverse solutions. Pointer Networks, which can satisfy the constraint that output sequences must be combinations of input sequence elements, have been widely applied in combinatorial optimization since their introduction.

Similar to the sequence-to-sequence \cite{sutskever2014sequence} network structure, Pointer Networks \cite{vinyals2015pointer} can satisfy constraints where the target output sequence must be a permutation or subset of the input source sequence. The Pointer Network is an encoder-decoder network structure, where both encoder and decoder are composed of Long Short-Term Memory (LSTM) units. Here, the Pointer Network's input is a sequence, and its output is a permutation combination of the sequence elements.


The "general" approach from Luong Attention \cite{luong2015effective} is adopted to calculate weights. For each target hidden state $h_{t}$, the attention score between $h_{t}$ and all source hidden states $h_{s}$, is calculated as follows:

\begin{equation}
	\text{score}(h_t, \overline{h}_s) = h_t^\top W_a \overline{h}_s
\end{equation}

where $W_a$ is a learnable weight matrix. The attention weights are then calculated using the softmax function, normalizing scores across all source hidden states:

\begin{equation}
	\kappa_{t}(s)= \frac{\exp(\text{score}(h_t, h_s))}{\sum_{s'} \exp(\text{score}(h_t, \overline{h}_{s'}))}
\end{equation}
	
A weighted sum of the source hidden states is performed, with weights given by the attention weights. This yields the context vector:

\begin{equation}
	c_t = \sum_s \kappa_{t} h_s
\end{equation}

The final output vector is a combination function of the context vector and current target state:

\begin{equation}
	\hat{h}_t = \tanh(W_c [c_t; h_t])
\end{equation}
where $W_{c}$ is a learnable matrix parameter.

\section{Experiments}\label{sec:experiments}
A typical example of multi-objective permutation-based combinatorial optimization is the Multi-Objective Traveling Salesman Problem. The single-objective Traveling Salesman Problem (TSP) requires visiting N cities exactly once and returning to the starting city while minimizing a single objective (such as the total distance between consecutive cities). The Multi-Objective Traveling Salesman Problem (MOTSP) aims to simultaneously minimize two or more potentially conflicting objectives. These distances can represent different metrics; for example, in a bi-objective problem, the first objective could be the sum of Euclidean distances between consecutive cities, while the second objective could be the sum of route risk costs. For a K-objective traveling salesman problem, it can be described as follows:

\begin{equation}
	min f(\tau ) = (f_{1}(\tau),f_{2}(\tau),\dots ,f_{K}(\tau))
	\label{eq:mtsp1}
\end{equation}
where $f_{k}(\tau)$ is calculated as:

\begin{equation}
	f_{k}(\tau)=\sum_{i=1}^{N-1} d^{k}_{\tau_{i},\tau_{i+1}}+d^{k}_{\tau_{N},\tau_{1}}
	\label{eq:mtsp2}
\end{equation}

where $\tau$ represents a permutation in the permutation space of ${1,2,\dots ,N}$. In this study, we set the number of optimization objectives to 2, with randomly generated symmetric distance matrices $D_{1}$ and $D_{2}$, where each element is randomly selected from the interval (0,1).

The Quadratic Assignment Problem (QAP) involves assigning $N$ facilities to $N$ locations to minimize the sum of the products of distances and flows \cite{knowles2003instance}.

\begin{equation}
	f(\tau)=\sum_{i=1}^{N}\sum_{j=1}^{N}a_{i,j}b_{\tau_{i}\tau_{j}} 
	\label{eq:qap}
\end{equation}
where $a_{i,j}$ represents the distance between locations $i$ and $j$, $b_{u,v}$ represents the flow between facilities $u$ and $v$. $\tau$ represents a permutation in the permutation space composed of ${1,2,\dots ,N}$.

\begin{figure}[htbp]
	\centering	
	\includegraphics[width=2.8in]{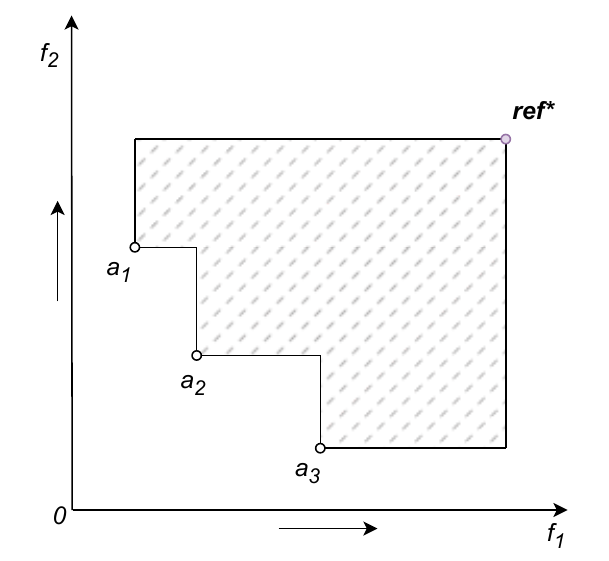} 
	\centering
	\caption{Hypervolume indicator}
	\label{fig:hypervol}
\end{figure}

The $K$-dimensional Multiobjective Quadratic Assignment Problem (MOQAP) aims to minimize the following $K$ functions simultaneously:
\begin{equation}
	min f(\tau) = (f_{1}(\tau),f_{2}(\tau),\dots,f_{K}(\tau))
	\label{eq:eemoqap1}
\end{equation}
where $f_{k}(\tau)$ is calculated as follows:

\begin{equation}
	f_{k}(\tau)=\sum_{i=1}^{N}\sum_{j=1}^{N}a_{i,j}b^{k}_{\tau_{i}\tau_{j}} 
	\label{eq:moqap}
\end{equation}
$b^{k}_{u,v}$ represents the $k$-th flow metric between facilities $u$ and $v$.

The Hypervolume (HV) metric \cite{zitzler2003performance} is used to evaluate algorithm performance. The Hypervolume indicator is widely used to compare different multiobjective optimization algorithms. Given a reference point $\mathbf{ref^{*}}$ in the objective space that is dominated by all solution sets' objective vectors, and $\mathbf{A}$ being the approximate Pareto front obtained in the objective space, the Hypervolume is defined as the area or volume of the region dominated by $\mathbf{A}$ and dominating $\mathbf{ref^{*}}$. The Hypervolume can be defined as follows:

\begin{equation}
	\left\{\begin{matrix}
		\mathbf{D} =  \{\mathbf{A'} \mid \mathbf{A} \prec \mathbf{A'} \prec \mathbf{ref^{*}};\mathbf{A}, \mathbf{A'},\mathbf{ref^{*}} \in \mathbf{R^{m}}\}\\
		Hypervolume(\mathbf{A}) = Volume(\mathbf{D})\\
	\end{matrix}\right.
	\label{eq:hv}
\end{equation}
where $\mathbf{A} \prec \mathbf{A'}$ indicates that $\mathbf{A}$ dominates $\mathbf{A'}$. $Volume(\mathbf{D})$ represents the area or volume of region $\mathbf{D}$. Taking bi-objective optimization as an example, for an approximate Pareto solution set $\mathbf{A}=\{\mathbf{a_{1}},\mathbf{a_{2}},\mathbf{a_{3}}\}$, as shown in Figure \ref{fig:hypervol}, the value of Hypervolume($\mathbf{A}$) represents the area of the shaded region.

If a solution set $\mathbf{A_{1}}$ dominates another solution set $\mathbf{A_{2}}$, then Hypervolume($\mathbf{A_{1}}$) $ >$ Hypervolume($\mathbf{A_{2}}$). When an algorithm obtains the true Pareto solution set, its Hypervolume value reaches the maximum. The larger the Hypervolume value of the approximate Pareto solution set obtained by a multiobjective optimization algorithm, the better the algorithm performs.

In this study, we compared the proposed SeqEMO and SeqEAG methods with NSGAII\cite{deb2002fast}, MOEA/D\cite{zhang2007moea}, and EAG-MOEA/D\cite{cai2014external} on the Multi-Objective Traveling Salesman Problem and Multi-Objective Quadratic Assignment Problem datasets. SeqEMO and SeqEAG are the proposed methods, which were obtained by training the population of MOEA/D and EAG-MOEA/D, respectively. The algorithms were fairly compared after 50,000 evaluations. The pointer network BatchSize was 128, learnRate was 0.001, Epochs was 200, numHiddenUnits was 200, and dropout was 0.05. As shown in Table \ref{tab:MOTSP}, except for the MOTSP30 dataset, SeqEMO outperformed MOEA/D on the other four MOTSP15, MOTSP20, MOTSP25, and MOTSP35 datasets. For EAG-MOEA/D, SeqEAG outperformed EAG-MOEA/D on two datasets and outperformed NSGAII on four datasets. From the MOQAP dataset in Table \ref{tab:MOQAP}, SeqEMO outperformed MOEA/D on all datasets. For EAG-MOEA/D, SeqEAG outperformed EAG-MOEA/D on the MOQAP25 and MOQAP30 datasets. The proposed methods can improve the performance of the original algorithms to some extent. It is worth considering comparing and verifying the algorithm's performance with more evaluation times.

\begin{table}[htbp]
    \renewcommand{\arraystretch}{2.0} 
    \footnotesize
	\caption{Comparison results on MOTSP instances.}
    \label{tab:MOTSP}
    \resizebox{1\textwidth}{!}{
    \begin{tabular}{l|c|c|c|c|c}
        \toprule
		&NSGAII              &MOEA/D                &SeqEMO              & EAG-MOEA/D           &SeqEAG \\
        \midrule	
		MOTSP15  &7.7289e-1 (1.38e-2) &7.5730e-1 (1.73e-2)   & 7.6796e-1 (1.15e-2)  &7.2803e-1 (4.67e-3) &7.2136e-1 (8.08e-3)\\
		MOTSP20  &7.3152e-1 (1.06e-2) &7.1300e-1 (1.26e-2)   & 7.1828e-1 (3.93e-3)  &7.2523e-1 (1.22e-2)  &7.3365e-1 (1.45e-2)\\
		MOTSP25  &7.5490e-1 (1.14e-2) &7.4576e-1 (1.79e-2)   & 7.5655e-1 (1.21e-2)  &7.6943e-1 (1.16e-2) &7.6532e-1 (8.25e-3)\\
		MOTSP30  &7.4402e-1 (6.65e-3) &7.4877e-1 (1.03e-2)  & 7.4077e-1 (1.21e-2)  &7.7705e-1 (4.74e-3)  &7.7237e-1 (7.40e-3)\\
		MOTSP35  &7.4219e-1 (9.53e-3) &7.5712e-1 (6.46e-3)  & 7.4730e-1 (1.13e-2)  & 7.9042e-1 (8.15e-3) &7.9362e-1 (1.01e-2)\\
        \bottomrule
        \end{tabular}}
\end{table}

\begin{table}[bt]
    \renewcommand{\arraystretch}{2.0} 
    \footnotesize
	\caption{Comparison results on MOQAP instances.}
    \label{tab:MOQAP}
    \resizebox{1\textwidth}{!}{
    \begin{tabular}{l|c|c|c|c|c}
        \toprule
		&NSGAII              &MOEA/D                &SeqEMO             & EAG-MOEA/D           &SeqEAG \\
		\hline	
		MOQAP15  &7.0535e-1 (3.06e-3) &6.9864e-1 (3.44e-3) &7.0326e-1 (1.44e-3)  &7.0426e-1 (2.49e-3) &7.0307e-1 (2.48e-3)\\
		\hline
		MOQAP20  &6.2601e-1 (1.57e-3) &6.2406e-1 (1.31e-3)  & 6.2625e-1 (1.61e-3)  &6.2459e-1 (1.93e-3) &6.2458e-1 (2.17e-3)\\
		\hline
		MOQAP25  &6.4208e-1 (1.00e-3) &6.4122e-1 (2.72e-3)  & 6.4247e-1 (1.82e-3)  &6.4197e-1 (3.84e-4) &6.4374e-1 (1.37e-3)\\
		\hline
		MOQAP30  &6.2697e-1 (1.10e-3) &6.2611e-1 (1.31e-3)  & 6.2699e-1 (1.02e-3)  &6.2769e-1 (1.60e-3) &6.2674e-1 (1.37e-2)\\
		\hline
		MOQAP35  &6.2623e-1 (1.06e-3) &6.2603e-1 (1.86e-3)  & 6.2641e-1 (8.70e-4)   & 6.2809e-1 (1.63e-3) &6.2848e-1 (8.63e-4)\\
		\bottomrule		
        \end{tabular}}
\end{table}

\begin{table}[htbp]
    \renewcommand{\arraystretch}{1.2} 
    \footnotesize
	\caption{Number of Predicted Solutions Updated}
    \label{tab:updatetimes}
    \resizebox{0.6\textwidth}{!}{
		\begin{tabular}{c|c|c|c|c|c|c}
			\midrule
			\midrule
			$i$-th Iteration&1th   &6th  &12th   &18th  &24th   &30th \\
			\hline	
			Updated Times &525   &286&106 & 113   & 50  & 80\\
			\midrule
			\midrule
			$i$-th Iteration&36th   &42th  &48th   &54th  &60th   &66th \\
			\hline
			Updated Times &112   &124&84&  68       & 105  & 45\\
			\midrule
			\midrule	
		\end{tabular}}
\end{table}

\begin{figure*}[htbp]
	\centering

	\begin{minipage}{0.4\textwidth}
		\centering
		\includegraphics[width=1.0\linewidth]{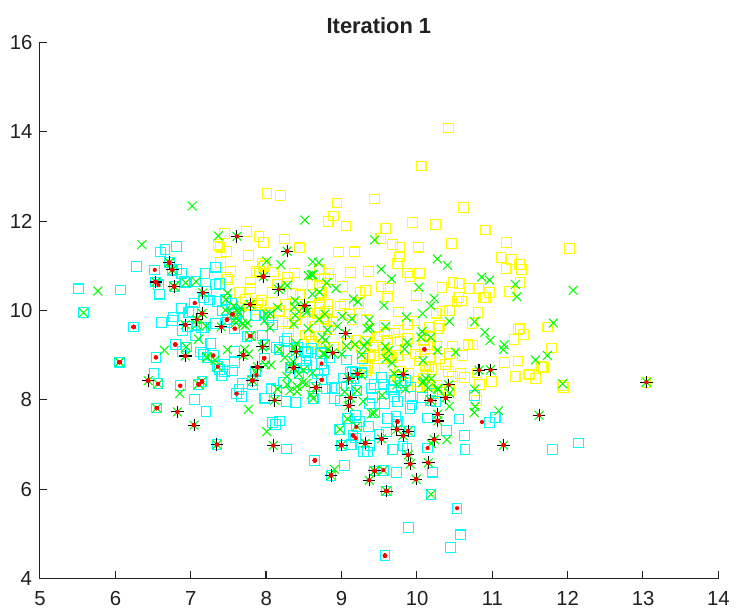}
		\subcaption{Iteration1}
	\end{minipage}
	\hfill
	\begin{minipage}{0.4\textwidth}
		\centering
		\includegraphics[width=1.0\linewidth]{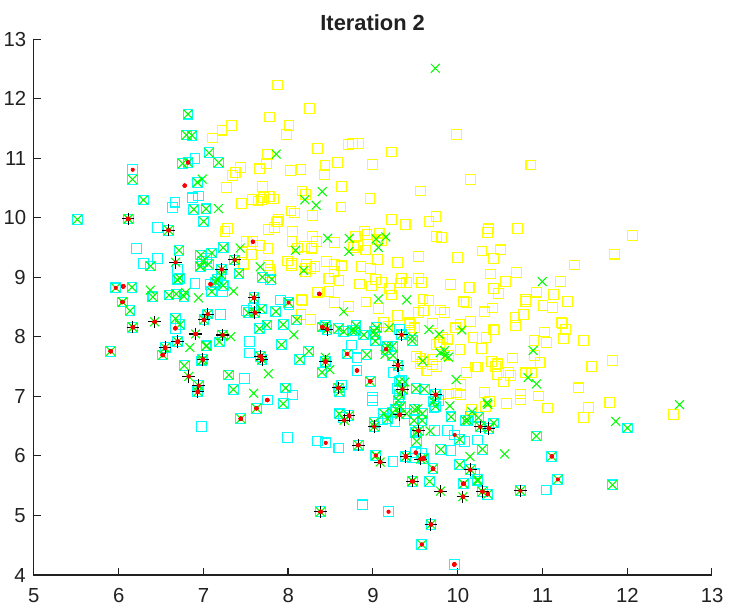}
		\subcaption{Iteration2}
	\end{minipage}
	
	\vspace{0.1cm} 
	
	\begin{minipage}{0.4\textwidth}
		\centering
		\includegraphics[width=\linewidth]{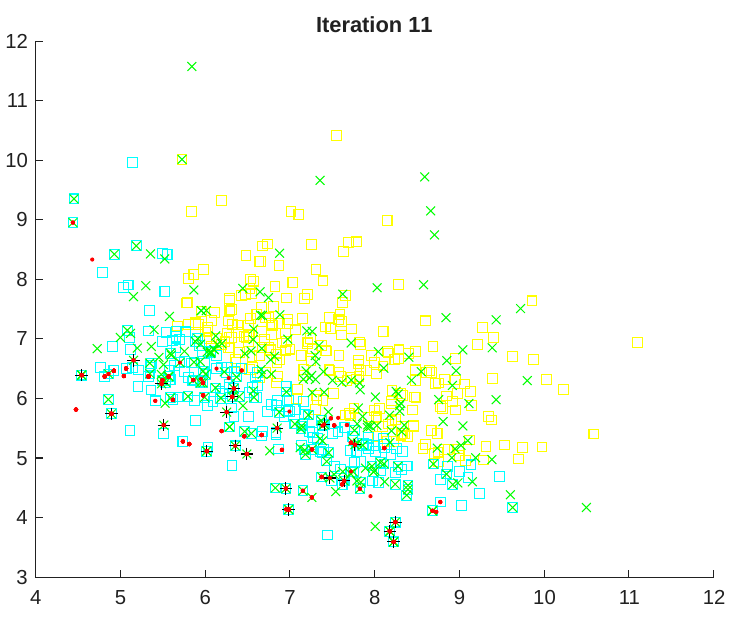}
		\subcaption{Iteration11}
	\end{minipage}
	\hfill
	\begin{minipage}{0.4\textwidth}
		\centering
		\includegraphics[width=\linewidth]{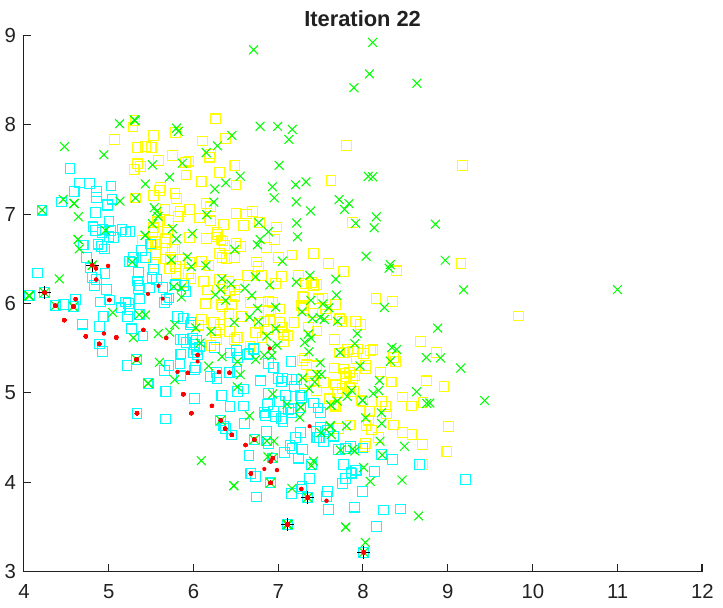}
		\subcaption{Iteration22}
	\end{minipage}
	
	\vspace{0.1cm} 
	
	\begin{minipage}{0.4\textwidth}
		\centering
		\includegraphics[width=\linewidth]{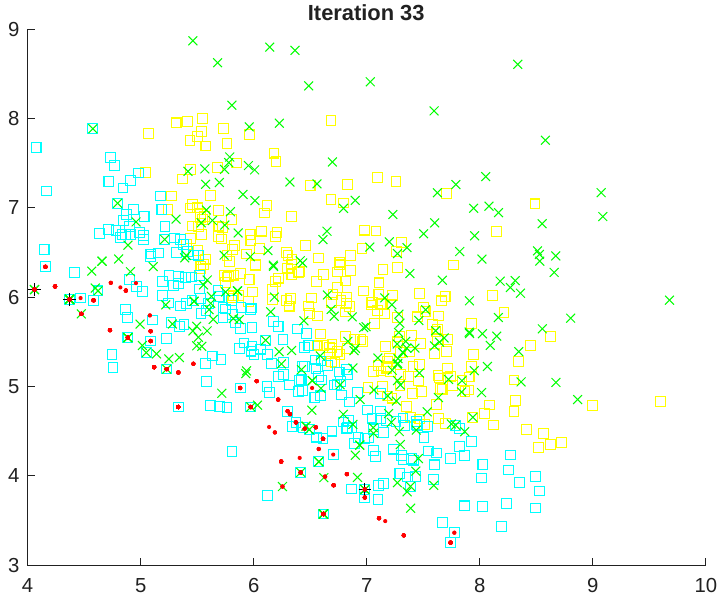}
		\subcaption{Iteration33}
	\end{minipage}
	\hfill
	\begin{minipage}{0.4\textwidth}
		\centering
		\includegraphics[width=\linewidth]{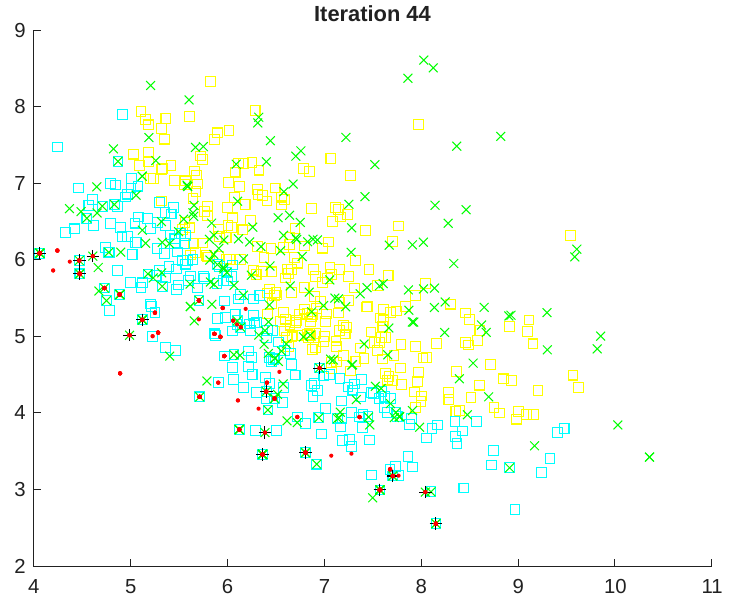}
		\subcaption{Iteration44}
	\end{minipage}
	
	\vspace{0.1cm} 
	
	\begin{minipage}{0.4\textwidth}
		\centering
		\includegraphics[width=\linewidth]{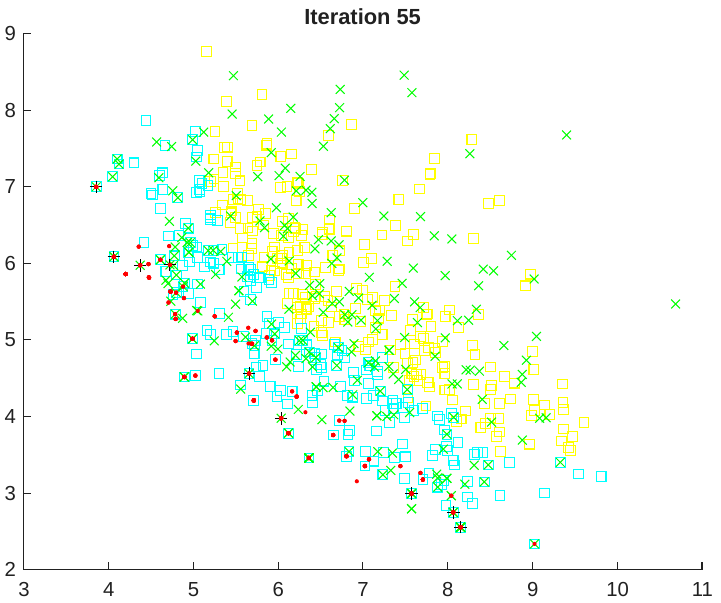}
		\subcaption{Iteration55}
	\end{minipage}
	\hfill
	\begin{minipage}{0.4\textwidth}
		\centering
		\includegraphics[width=\linewidth]{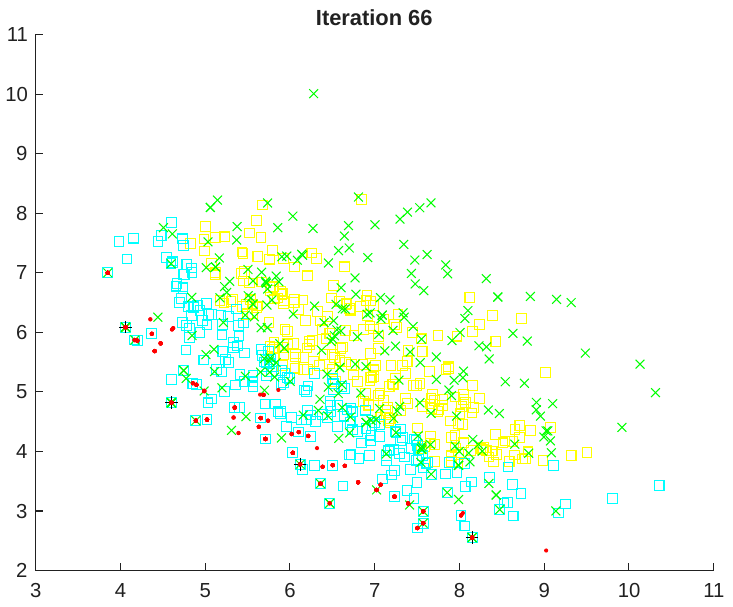}
		\subcaption{Iteration66}
	\end{minipage}	
	
	\caption{Distribution of Objective Value of the Predicted Solutions}
	\label{fig:ptusefual}			
\end{figure*}

To validate the contribution of solutions generated during training and prediction to the population, Table \ref{tab:updatetimes} lists the number of individuals (times) in the population that were effectively updated by solutions generated in the i-th iteration, where "1th" represents the first iteration. It can be observed that a large number of solutions in the population were updated by the generated solutions, demonstrating the effectiveness of the pointer network's training and prediction process. Although the number of updates showed a decreasing trend with increasing iterations, even in the final iteration, 45 solutions were still updated by the generated solutions, indicating that the neural network prediction could continuously generate valuable solutions.

Additionally, visualization methods were employed to display the population migration process through pointer network predictions. As shown in Figure \ref{fig:ptusefual}, the distribution of the population in the objective space is plotted every 11 iterations. Yellow squares "$\Box$" represent inferior solutions in the offsprings, blue squares "$\Box$" represent superior solutions, red dots "$\cdot$" represent the updated population, green "x" marks represent solutions generated from inferior offsprings predictions, and "+" marks represent population solutions updated by predictions from inferior offsprings solutions. The figure clearly shows the migration process where yellow inferior solutions are transformed into green solutions, with an evident overall migration toward the Pareto front. Furthermore, many of the migrated solutions do not overlap with the previous superior solutions in the offsprings but are distributed nearby, enhancing the diversity of offsprings solutions. This results in the predicted offsprings being distributed close to the approximate Pareto front while exploring more possibilities.

\begin{figure*}[htbp]
	\centering	
	\begin{minipage}{0.4\textwidth}
		\centering
		\includegraphics[width=0.95\linewidth]{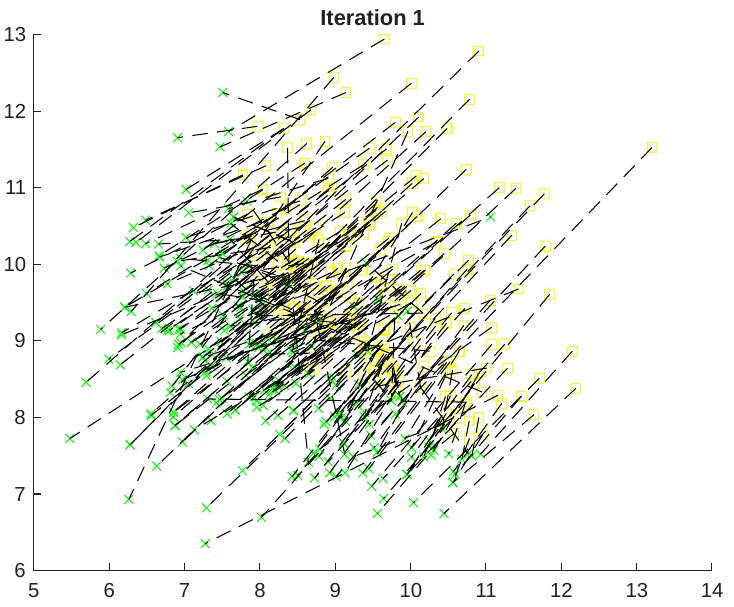}
		\subcaption{Iteration1}
	\end{minipage}
	\hfill
	\begin{minipage}{0.4\textwidth}
		\centering
		\includegraphics[width=0.95\linewidth]{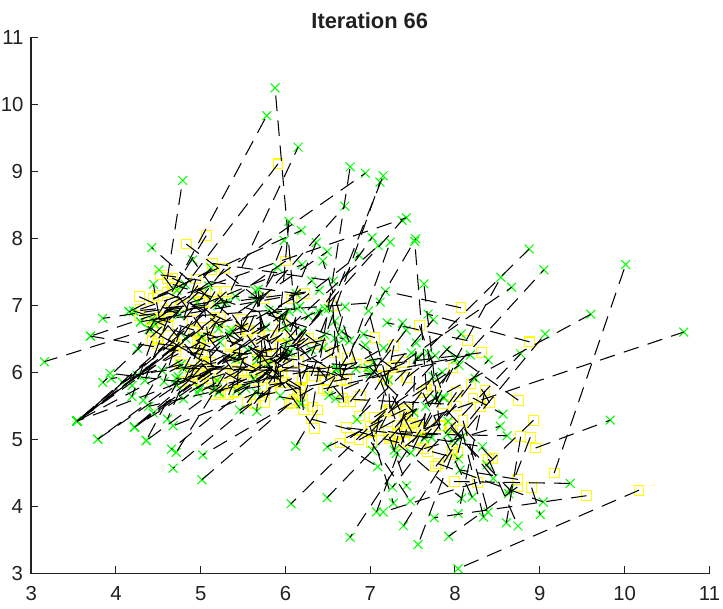}
		\subcaption{Iteration66}
	\end{minipage}
	\vspace{0.1cm} 
	\caption{Migration Direction of Predicted Approximate Solution Sets in Objective Space}
	\label{fig:MAP}	
\end{figure*}

Moreover, as shown in Figure \ref{fig:MAP}, the one-to-one correspondence between the inferior population and the predicted generated population is displayed. Black dashed lines indicate the migration process from each inferior solution in the offsprings to newly generated solutions. It is evident that the vast majority of migrations occur in an orderly manner toward the coordinate origin, continuously approaching the Pareto front direction, which demonstrates the algorithm's effectiveness.

\section{Conclusion}\label{sec:conclusion}
To fully utilize the large amount of data generated during the multi-objective optimization algorithm's iteration process, we proposed a learnable evolutionary multi-objective combinatorial optimization method based on a sequence-to-sequence model, combining the advantages of neural networks and multi-objective evolutionary algorithms. The method divides each generated sub-population into elite and poor solutions, creating training data and labels through one-to-one matching based on solution angles in objective space. Through training on these data, newly predicted solutions can effectively migrate original solutions toward the Pareto front direction, increasing the number and diversity of high-quality solutions in the population. This effectively enhances the algorithm's exploration capability and improves its performance, demonstrating the generation capability of the sequence-to-sequence model-based Pointer Network in effectively increasing the diversity of high-quality solutions in the population.

\bibliography{main}

\end{document}